# MIDAS – An Influence Diagram for Management of Mildew in Winter Wheat

Allan Leck Jensen
Dina (Danish Informatics Network
in the Agricultural Sciences)
Research Centre Foulum
P.O. Box 39, DK-8830 Tjele, Denmark
alj@dina.sp.dk

Finn Verner Jensen
Department of Computer Science
Aalborg University
Fredrik Bajers Vej 7E,
DK-9220 Aalborg Ø, Denmark
fvj@iesd.auc.dk

## Abstract

We present a prototype of a decision support system for management of the fungal disease powdery mildew in winter wheat. The prototype is based on an influence diagram which is used to determine the optimal time and dose of mildew treatments. This involves multiple decision opportunities over time, stochasticity, inaccurate information and incomplete knowledge. The paper describes the practical and theoretical problems encountered during the construction of the influence diagram, and also the experience with the prototype.

## 1 INTRODUCTION

In Denmark it is a widespread opinion, publically as well as politically, that the environmental impact of agricultural production must be reduced. Findings of pesticides and nitrogen residues in drinking water have induced the government to introduce action plans for a significant reduction of the consumption of fertilizers and pesticides.

Applied unwisely, reductions in the agricultural input factors, like fertilizers and pesticides, involve risks of inadequate effects and hence economical losses. On the other hand, input factors are expensive (and farmers are generally known to be thrifty), so the main reason for farmers to apply excessive amounts of input factors is to secure themselves against the risks of inadequate effects of the input factors. They will be eager to reduce the consumption of input factors and save money, if they are given access to robust recommendations of when it is safe to reduce the doses. These recommendations could come from decision support systems. Hence, a change in the farmers decision policy from *insurance farming* to *precision farming* should only be expected in the pace of the development and acceptance of such systems.

A few computer-based decision support systems for crop protection have reached an operational level already. For example, the Danish system PC-Plant Protection (Murali, 1991; Secher, 1991) gives case specific recommendations based on empirical threshold values of diseases and pests.

Generally, such decision support systems recommend significantly lower doses than the recommended dose of the chemical products, the so-called label dose. However, theoretically there is still room for further reductions of the recommendations from these conventional systems: First of all, they can be expected to recommend too high doses in general, because the only way they can handle the uncertainty of the domain is by being cautious. By demanding frequent field observations they can relax this cautiousness.

Secondly, both empirical and deterministic decision support systems postpone treatment recommendations until a certain threshold value (empirical or calculated) has been reached, where the expected damage exceeds the cost of treatment. For these cases, earlier treatments could have solved the problems of disease or pest incidence with lower pesticide doses, had it been known that they would develop into problems.

Thirdly, there may be cases where the conventional systems overestimate the expected damage and recommend treatments which could be avoided, had it been known that the disease or pest incidence would not develop into problems after all.

Hence, in theory there may be cases where it will be more beneficial to recommend a treatment earlier than the conventional systems, and other cases where it will be more beneficial to postpone the treatment and keep the development under observation.

Here, a prototype of a decision support system for management of the fungal disease mildew in winter wheat will be presented. It has been the challenge in the construction of the prototype to narrow the gap between 'too early safety treatments' and 'too late emergency treatments' by handling the naturally encumbered uncertainty formally in an influence diagram. The decision support system uses the influence diagram to give case specific recommendations of timing and dosage of mildew treatments. The name of the decision support system, MIDAS[1], is an acronym for Mildew Influence Diagram for Advice of Sprayings. MIDAS has been constructed as a Ph.D. project (Jensen, 1995a). The Ph.D. thesis can be down-loaded (*http://www.sp.dk/~abj/*) entirely or in parts.

## 2  THE DECISION PROBLEM

Powdery mildew (*Erysiphe graminis*) is a fungal disease which subsists on living plant tissue, mainly leaves. Infection is mainly by wind spread spores. The infected plant cells will first show reduced photosynthesis and increased respiration, and later they will die. The disease development is highly influenced by weather conditions, mainly temperature, humidity and wind. Under weather conditions which are favourable for the disease, it may spread rapidly, while under unfavourable conditions it will not spread, and the present disease may disappear with time due to the emergence of new, uninfected leaves and the death of old, infected leaves.

When a farmer observes mildew in his winter wheat field, he is facing a decision problem. Based on his field observations and expectations to the future he will have to determine the optimal treatment decision for the current disease problem.

In order to give appropriate recommendations, MIDAS is intended to mimic the decision context of the farmer. Therefore MIDAS contains an influence diagram model of the decision context, i.e. the influence diagram represents the treatment decisions, the relevant information, which the decisions are based upon, and variables forecasting the consequences of the decisions to the crop and the disease. This influence diagram model in MIDAS is called the *decision model*.

The decision model is a forecasting model describing the development of crop and disease from the time of enquiry to crop maturity. Being an influence diagram, it can also be used to determine the optimal decision. The decision optimization is based upon relevant information about the crop, the disease and the weather. These types of information are described in table 1.

In order to model the decision context of the farmer properly, the decision model of MIDAS contains a sequence of treatment decisions, covering the remaining part of the growing season, rather than only the decision for the current treatment. The reason for this is, that the current decision problem can not be solved optimally, if it is considered isolated from future de-cisions and information. For example, a decision to refrain from spraying now will seldomly be considered optimal, if the possibility of spraying in the future is not taken into account. Therefore, MIDAS determines which decision alternative for the current treatment decision is the optimal, under the assumption that all future decisions will be made optimally according to the available information at the time.

---

[1] Coincidentally, Midas is also the name of a mythical king of Phrygia, who was given the ability that everything he touched turned into gold. This gave King Midas severe digestional problems! If it should be attempted to extract a moral from this story in the context of fungicide sprayings in wheat, it could be that farmers should remember to treat wheat not only as a source of gold but also as a source of food.

Table 1: Variable types in MIDAS. *Static* information variables are variables where the value is considered constant and known before the time of the first treatment opportunity. The value of *dynamic* information variables change with time, so current values are recorded before the decision optimization.

| Variable type | Variable |
|---|---|
| Static information | Winter wheat variety, Nitrogen fertilization strategy, Soil type, Plant density |
| Dynamic information | Weather, Disease incidence, Remaining time to harvest, Cost of fungicide, Label dose of fungicide, Cost of spraying, Expected grain yield, Expected price of grain |
| Decisions | Dose of treatment (0 possible) |
| Utilities | Value of yield, Cost of treatment, Value of disease induced yield loss |

A typical recommendation from MIDAS could be to wait for some days and reconsider the treatment decision after a new observation of the disease level at that time. This recommendation corresponds to the decision alternative of the current treatment decision to apply no fungicide treatment. This treatment decision would be optimal, if the risk of an epidemic outbreak during the period was considered to be sufficiently low.

Obviously, the decision optimization in MIDAS is affected by uncertainty of several sources.

1. Stochasticity. The weather and the disease infections are factors influencing the growth of crop and disease with an element of unpredictability.

2. Inaccurate observations. The problem is mainly the field recordings of disease level which are difficult and error-prone.

3. Incomplete knowledge. The domain contains relations which can be considered deterministic in

principle, but since the relations are not known exactly, the interpretation of them in the decision models involves uncertainty.

This uncertainty is handled by using influence diagrams.

## 3  TIME

The decision model is a dynamic influence diagram, consisting of a sequence of time steps, which covers the remaining part of the growing season. Each time step contains a single treatment decision variable and information variables to describe the state of the domain at the particular time, as described in table 1. Each time step also contains chance variables to describe important, latent variables. The relations of a time step describe the current state of the biological system and the development during the time span of the time step until the beginning of the next time step. This makes time an important parameter in the decision model.

Chronological time is not sufficient to model the development of most biological processes, since they are normally also influenced by the weather, and especially by temperature. Therefore, each time step should be able to describe the influence of weather on the development during the time step. However, the best you can say about the weather more than a week from now is invariant over the years. Therefore, a forecasting model can only have prior probabilities on the weather nodes, and they can just as well be avoided by including the expectations to the weather in the other conditional probabilities.

In several cases the development also depends on the physiological characteristics of the crop, as described by the crop development stage. For example, the prediction of the disease level in the next time step depends on the crop development stage, because the development stage influences the rate of death of old (potentially infected) leaves and the emergence of new leaves, and the disease level is defined as the fraction of the green leaf area with disease symptoms.

Ideally, the decision model could be constructed with time steps of a fixed length in chronological time. For practical reasons this fixed length should be an integral number of days, and preferably one week, as this would enable the farmer to fit the tasks of field observations and decision making concerning fungicide treatments more easily into his routines.

In this case the decision model would have to represent the influence of crop development stage and weather on the development of the biological processes. This was the approach of early versions of the decision model, but the complexity costs of representing all three influentials on the development (i.e. chronological time, temperature and crop development stage) turned out to be too large.

Therefore, the three influentials on development were integrated into a single developmental time scale, the *thermal time scale*. The thermal time scale was defined as the temperature sum, i.e. the accumulated daily mean air temperature, above a base temperature of 5°C. This thermal scale was assumed to be a good time measure for the description of the different biological processes, like crop growth, crop development, disease growth and fungicide decay.

The thermal time scale was defined to be the expected temperature sum remaining to crop maturity. This remaining thermal time should be divided into thermal time periods, each corresponding to a time step of the decision model. In order to approach the advantages of a chronological time scale, it was decided to give these thermal time periods an expected length of one chronological week. So the length of a time step is called a *thermal week*. The actual length of a time step in chronological time depends on the temperature, but with average temperature it will be one week. In the current version of MIDAS historical climate data have been used to determine the duration in thermal time of the time steps, but actually the duration of the first time step could be determined from the weather forecast.

In order to calibrate the thermal time of the decision model the farmer is asked to give his estimate of the number of weeks to crop maturity.

## 4  THE INITIAL INFLUENCE DIAGRAM

One possible strategy for the representation of the system would be to store a compiled, general decision model. At the time of enquiry, relevant information about the current state and the history of the domain could then be used to calibrate the general decision model to the specific case. However, a representationally less complex strategy was chosen, where a case-specific decision model is constructed at each time of enquiry.

As illustrated in figure 1 the decision model is assembled from a sequence of case-specific time step modules, each of which represents a remaining time step. The case-specific time step modules are all processed from a general module called the *general decision model module*, abbreviated to the *GDM module*. During the assemblage of time step modules, utility variables are represented. The utility variables of a time step are the cost of the treatment and the value of the disease induced yield loss of the time step. They are calculated from information provided by the farmer. The assembled, case-specific decision model will have to be compiled before it can be used for reasoning and decision optimization.

The GDM module is displayed in figure 2 in a representation where all variables are rectangular. All the variables are chance variables, except the one labelled Treatment_1, which is a decision variable.

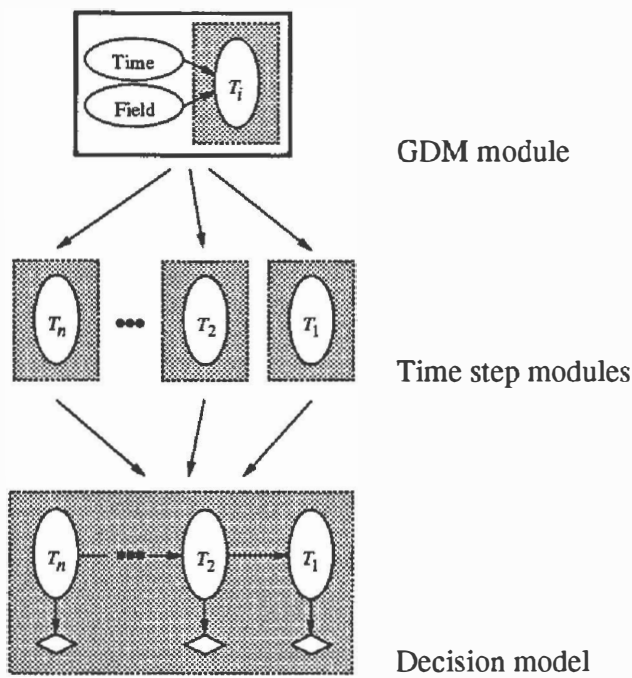

Figure 1: Procedure for decision model assemblage. By entering information about the particular time and field, the GDM module is made case-specific. The proper sequence of case-specific time step modules is generated and assembled into the case-specific decision model. Note that the time steps are counted down from $n$ to 1.

The GDM module is general in the sense, that it describes any time step of the season for different selections of cultivation factors. The GDM module is made case-specific by setting the variables on the left hand side of the frame in figure 2: TimeStep defines the current thermal week on the thermal time scale, while the other variables characterize the cultivation factors, i.e. sowing density, soil type, nitrogen fertilization and crop variety.

Each time step of the decision model contains an instance of each of the variables inside the frame of figure 2, and in order to distinguish the different instances, the variables are given an index number corresponding to the number of the time step they belong to. In the GDM module the number 1 is used arbitrarily for the index of the variables within the frame. Consequently, the three variables on the right hand side of the frame, representing variables of the next time step, have been given the index number 0.

The GDM module contains nodes describing relevant attributes of the crop, the disease, the weather, the management, and the fungicide. The focal variable of the module is Treatment, which describes the dose of the treatment decision of the time step. The Disease-Level variables describe the proportion of the green leaf area which is covered by disease at different phases relative to the Treatment decision, as indicated by the extension B or A, corresponding to Before and After the Treatment, respectively. The DiseaseLevel is not observed directly, but it is calibrated by a simpler field observation of the proportion of plants with disease symptoms, DiseaseObserv. The development of disease is determined by an intrinsic GrowthRate. The GrowthRate is affected by weather (ClimateEffect), the density of the crop (CropStructure), and the general protection level against disease infection of the crop (MeanProtectn). MeanProtectn consists of a basic protection level (BasicProtection) determined by the resistance genes of the winter wheat variety and the nitrogen fertilization strategy, and of an additional contribution from protective chemicals of previous treatments (ProtectnLevel). ProtectnLevel is determined from the concentration of protective chemicals (ProtectnConc). The proportion of leaves which have emerged since the previous treatment (NewLeafFract) determines the relative effect on MeanProtectn of these protection levels on untreated and treated leaves, respectively. In order to determine the current NewLeafFract and ProtectnConc, the time and dose of the previous treatment are stored (PrevTreatment and PrevDose). YieldLossPct is the expected relative yield loss which is caused by accepting disease at the given level during the given time step.

It was not possible to quantify the GDM module directly from field experimental data, in spite of agriculture being an extremely data rich domain. The most important reason for this paradox is the distance between input and output data of field trials in terms of causal relations. Data from field trials typically describe the effect of combinations of cultivation factors on response variables, like yield, leaf area or disease progress, while the BN model may require intermediate explanatory variables and a dense network of direct cause-effect relations to combine the cultivation factor input with the response variable output of the trials. In this way, it has been surprising to learn that the causal dependency structure, reflected by the GDM module, is well understood by the experts, but there has been almost no attempts to quantify these cause-effect relations.

As a consequence of these quantification complications, simple deterministic models have been constructed and used to quantify the decision model (Jensen, 1995b). Each quantification model describes a child variable as a function of its parent variables. Together these quantification models build links between the well-documented concepts of the GDM module, across the latent parts.

For most variables the uncertainty associated to the quantification model has not been represented properly in the present version of the prototype. Uncertainty from stochasticity and from inaccurate observations have been better represented, though.

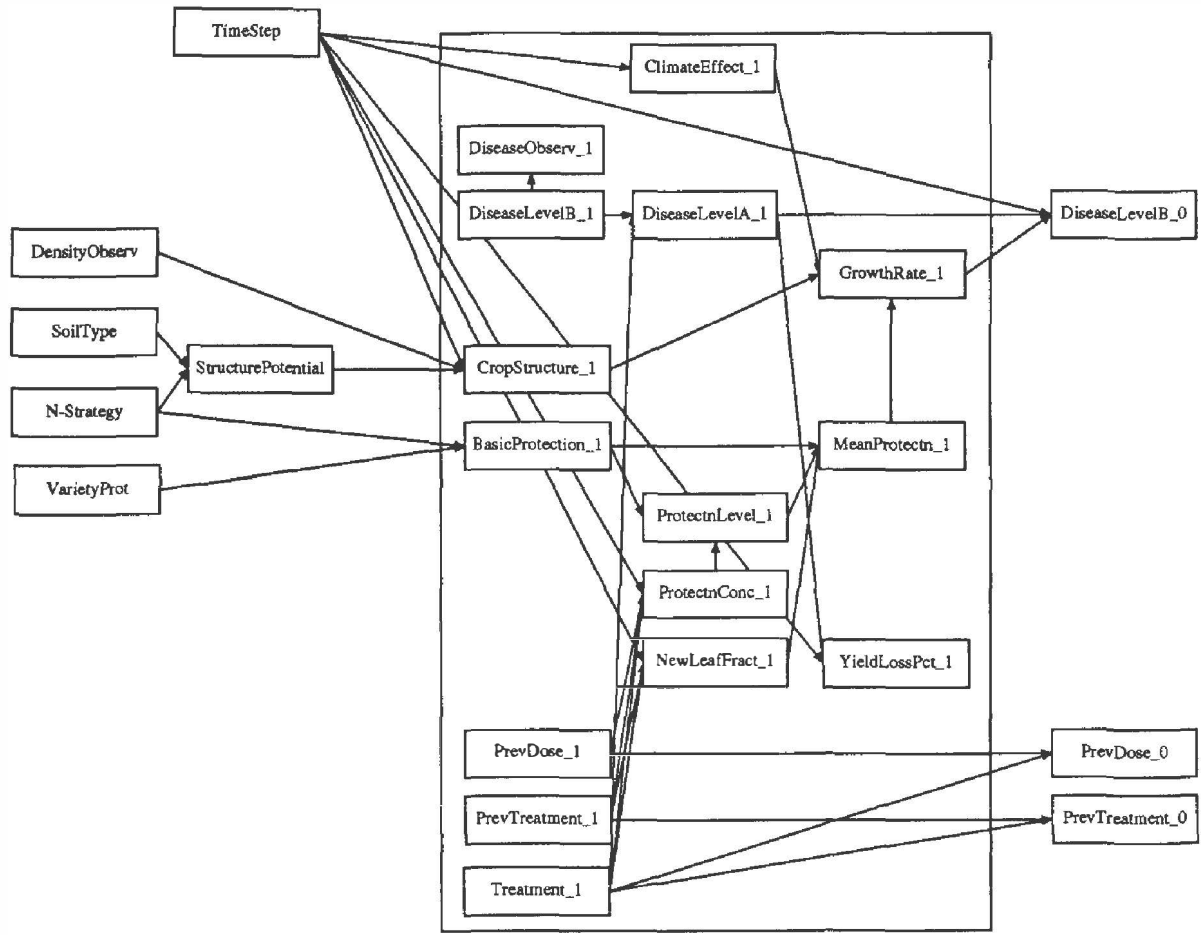

Figure 2: The GDM module. The variables are described in the text.

It would be possible to use the GDM module together with case-specific information to construct a specific Bayesian network for consequence simulation under uncertainty. The main use of the GDM module, however, was intended to be for construction of case-specific influence diagrams for decision optimization, however.

The decision optimization is performed with the algorithm implemented in Hugin by Jensen, Jensen and Dittmer (1994), which is based on more general schemes by Shenoy (1992) and Shachter and Peot (1992). Unfortunately, even though the junction tree corresponding to a Bayesian network version of the decision model was tractable, the specialized junction tree (Jensen et al., 1994) corresponding to the influence diagram version turned out to lead to the combinatorial explosion. As a consequence, it was necessary to modify the initial GDM module.

## 5 THE CURRENT INFLUENCE DIAGRAM

The combinatorial explosion for the initial influence diagram is caused by insufficient monitoring of the state of the crop and disease. To realize this, a brief description of the algorithm for decision optimization may be appropriate.

The algorithm follows the principle of dynamic programming (Bellman, 1957) and considers the decisions in the opposite order than they are made. First, the final decision is considered, and for each information scenario at that time, the decision alternative with optimal expected utility is determined. Second, the final-but-one decision is considered, and for each information scenario at the time of that decision, the decision alternative with optimal expected utility, under condition of optimal decision making for the final decision, is determined. Subsequently, the preceding decisions are considered in reverse order, and each of them is optimized under assumption of optimal decision making in the future.

The primary reason for the large computational com-plexity of the decision optimization for the initial influence diagram is connected to the term "each information scenario" used above. The set of information scenarios at the time of a decision consists of all configurations of observed variables which are d-connected to a utility node influenced by the decision. This set may become intractably large.

The problem with the initial GDM module is that the disease level, DiseaseLevel, is not observed directly, as explained earlier. Instead, the simple disease incidence measure, DiseaseObserv, is used to calibrate the DiseaseLevel variable, as shown in figure 2. This means that the current state of the disease depends on not only the current value of DiseaseObserv, but also all the previous, together with all the previous treatment decisions and other characteristics of the past.

In order to make the decision model tractable for decision optimization it was modified to fulfill the condition that the local information of the system overwrites all previous information. This condition is called the *information blocking* condition.

To be more specific, let the set of variables of the decision model be partitioned into succeeding time steps $T_i$ with decision variable $D_i \in T_i$, for $i = 1, \ldots, n$. Furthermore, let $I_1$ denote the set of observed variables before $D_1$, and $I_i$ the set of variables which are observed after $D_{i-1}$ and before $D_i$, for $i = 2, \ldots, n$.

The information blocking condition can be formulated in this notation as follows:

$$P(Y \mid I_k, D_k, X) = P(Y \mid I_k, D_k)$$
$$\text{for } X \in \bigcup_{i=1}^{k-1} T_i, \ Y \in \bigcup_{i=k+1}^{n} T_i \text{ and } k = 1, \ldots, n$$

This is a stronger condition than the Markov property, which is fulfilled by a decision model constructed from the initial GDM module, since also unobserved variables of the time step are included in the blocking of the Markov property:

$$P(Y \mid T_k, X) = P(Y \mid T_k)$$
$$\text{for } X \in \bigcup_{i=1}^{k-1} T_i, \ Y \in \bigcup_{i=k+1}^{n} T_i \text{ and } k = 1, \ldots, n$$

The left hand side of figure 3 shows the correct causal dependency structure of the relationships between DiseaseObserv and the DiseaseLevel nodes, while the right hand side shows the changed structure where the DiseaseObserv node blocks the influence from the past through DiseaseLevelB. The modified structure fulfills the information blocking condition, but it is incorrect. It was assumed, however, that the accuracy of the model predictions would not be deteriorated significantly.

An example of a decision model with three time steps constructed from the modified GDM module is shown in figure 4.

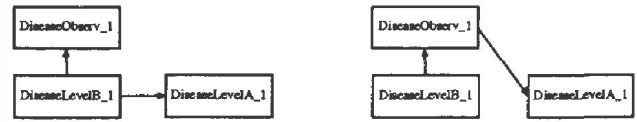

Figure 3: Left: The initial causal structure of the relationships between the DiseaseObserv and the DiseaseLevel nodes. Right: The modified structure to achieve a blocking of the past by the observed nodes (DiseaseObserv).

The structural change displayed in figure 3 made it necessary to change the conditional probability table for DiseaseLevelA_1 in the GDM module. This was done by applying simple probability calculus on the available conditional probability distributions.

## 6 EXPERIENCE

The experience from MIDAS can be described through the quantitative and the qualitative performance of the system. The representational and computational complexity of the influence diagram belongs to the quantitative performance of MIDAS. On the other hand, the correctness of the predictions of the influence diagram belongs to the qualitative performance.

To begin with the quantitative performance, the total number of conditional probabilities needed to specify the GDM module is 14,917. Naturally, this measure is not very descriptive of the complexity of the influence diagram – the total clique size of the junction tree of a decision model is more appropriate.

The total clique size of a decision model with a single time step is 27,388 probabilities, and it increases with 83,196 for each additional time step. This gives a maximum clique size of 1.6 million for a model with 20 time steps, which is considered to be the highest number of time steps for realistic applications. Without compression it requires 14.8 Mb to store this model.

Even though the times for creation and application of the decision model are machine dependent, the times listed in table 2, which were measured on a Sun SPARCstation ELC, can give an impression:

Table 2: Performance times by MIDAS.

| Task | Time |
|---|---|
| Assemblage | 1 sec |
| Compilation | 100 secs |
| Loading | 25 secs |
| Propagation | 25 secs |

The qualitative performance of MIDAS has been evaluated in a limited series of tests, where the predictions and recommendations of MIDAS were compared with real-life field data. The results of the tests were generally satisfactory, but they also revealed certain possible improvements.

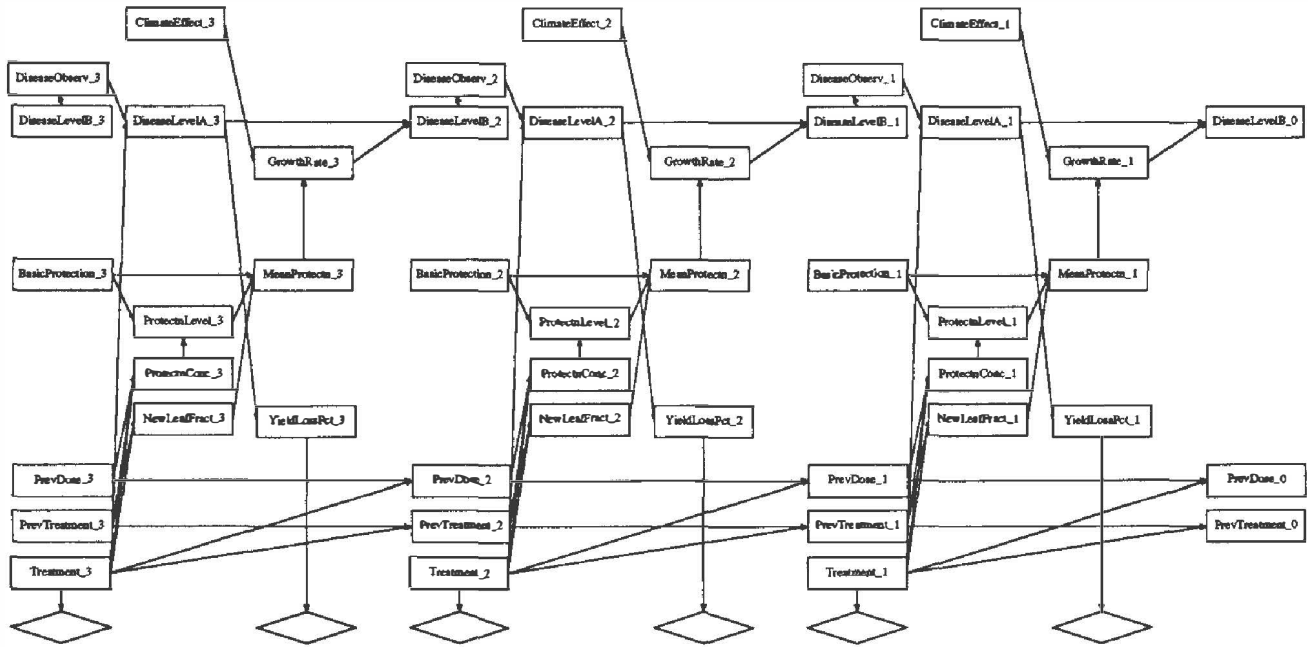

Figure 4: A decision model with three time steps. Nodes labelled Treatment_$i$ for $i = 1, 2, 3$, are decisions, while the other rectangular nodes are chance nodes.

One of the main subjects of the validation tests was an evaluation of the predictions of DiseaseLevel. This was done both with the current influence diagram with information blocking and with the initial influence diagram with the true causal structure of the DiseaseLevel and the DiseaseObserv variables. These two structures are referred to in the following as the *true* and the *approximate* structures, respectively.

Data sets from field experiments with frequent observations of disease level are rare, but a data set with 960 observations of DiseaseLevel was obtained. The observations were made weekly for 10 consecutive weeks and in 96 experimental plots. Unfortunately, the data set had only 154 observations of DiseaseObserv, and they were not randomly distributed on the plots, making it difficult to draw firm conclusions for the tests.

A comprehensive description of the validation tests is given in (Jensen, 1995a), but in summary some of the conclusions can be listed:

- When the information for calibration of the current disease level is available, the predictions of the model with the true structure are good. On average the mean value of the predicted DiseaseLevel distributions was 2.75 (s.d. 5.48), which is surprisingly close to the average of observed disease levels, 2.78 (s.d. 5.86).
- The model with approximate structure shows a less satisfactory behaviour. It tends to underestimate high disease levels and overestimate low disease levels. The reason for this is, that in order to let it resemble the true model the probability tables for the approximate model were calculated from the tables of the true model. However, the distribution of DiseaseLevelB, $P(DLB)$, was needed in the calculations, and a fixed prior distribution was used. Hence, the two structures only give the same result, when the marginal distribution of $DLB$, given the entered evidence, is equal to this estimated prior distribution.
- The predicted probability distributions are too narrow. The reason for this is the insufficient representation of uncertainty in the quantification models, as mentioned in section 4.
- The variable DiseaseObserv was intended to be a simple measure for calibration of the DiseaseLevel variable. For the model with approximate structure, however, the role of DiseaseObserv is much more crucial for the predictions and recommendations of the influence diagram, and the tests have shown that it is too simple for this purpose. The problem occurs, because when about 2% of the green leaf area is infected (DiseaseLevel $\approx$ 2%) then almost all plants are infected (DiseaseObserv $\approx$ 100%). This means, that if the value of DiseaseObserv is close to 100%, then the corresponding value of DiseaseLevel may be any number between about 2% and 100%.

# 7 FUTURE WORK

The experience from the validation tests of MIDAS has demonstrated several possible improvement of the system, as described in the previous section.

To improve the predictions of DiseaseLevel for the model with information blocking, several different prior distribution for $P(DLB)$ should be used in order to fit the actual situation better. A typical distribution of $DLB$ could be estimated for each of the 120 combinations of TimeStep and BasicProtection, and then these priors could be used to calculate a series of conditional probability tables for $P(DLA \mid DO, T)$. When the particular decision model is assembled, the proper version of $P(DLA \mid DO, T)$ can be selected for each time step. This would neither increase the representational nor the computational complexity of the decision model – only the representational complexity of the GDM module.

Obviously, it is an important task for future improvements of MIDAS to improve the representation of uncertainty in the GDM module, and most urgently the representation of uncertainty from incomplete knowledge associated with the quantification models needs to be improved.

In order to improve the calibration of DiseaseLevel, and hence the recommendations, an additional information node should be introduced in the GDM model. It is important for the operational use of the system, that a good trade-off is found between the value of the information and the complexity and time requirement of obtaining it. For example, if the farmer's rough estimate of the DiseaseLevel is informed when DiseaseObserv is close to 100%, it would be a simple way to improve the calibration significantly, even with only a few states of the variable, like "0–2%", "2–20%" and "20–100%".

A different kind of improvement would be to consider a relaxation of the information blocking condition. For example, the influence diagram could consist of time steps with alternating structures, say with only every second time step having the information blocking structure and the other time steps having the true causal structure.

It should be mentioned, that it will be possible to use both model structures in parallel: The approximative (with information blocking) for decision optimization and the true for reasoning (for example about the consequences of a given treatment).

The mentioned improvements would probably result in more accurate predictions of the system, and hopefully the predictions of the influence diagram with approximate structure would be comparable to those with the true structure.

Following such improvements, the next natural step would be to promote MIDAS further from a prototype to an operational recommendation system used by farmers and agricultural advisors. In order to do this, it would be necessary to gain confidence in the robustness of the recommendations by performing a thorough validation on field level. Even though much can be learned by validations from field data, there are limitations. For example, the validation tests showed that MIDAS generally recommends lower treatment doses than the empirical recommendation system PC-Plant Protection. This is not an impressing result, however, as long as the consequences of actually following the recommendations are not known.